\documentclass[sigconf, nonacm=true]{acmart}
\renewcommand\footnotetextcopyrightpermission[1]{} 
\usepackage{multirow}
\usepackage{subfigure}
\usepackage{algorithm}
\usepackage{algorithmic}
\usepackage{rotating}
\AtBeginDocument{%
  \providecommand\BibTeX{{%
    \normalfont B\kern-0.5em{\scshape i\kern-0.25em b}\kern-0.8em\TeX}}}

\setcopyright{acmcopyright}
\copyrightyear{xxxx}
\acmYear{xxxx}
\acmDOI{XXXXXXX.XXXXXXX}

\acmConference[Conference acronym 'XX]{Make sure to enter the correct
  conference title from your rights confirmation emai}{June 03--05,
  2018}{Woodstock, NY}
\acmPrice{15.00}
\acmISBN{978-1-4503-XXXX-X/18/06}

\begin{document}

\title{ASE: Anomaly Scoring Based Ensemble Learning for Highly Imbalanced Datasets}

\author{Xiayu Liang}
\email{lxy_double7@foxmail.com}
\affiliation{%
  \institution{South China University of Technology}
  \city{Guangzhou}
  \state{Guangdong}
  \country{China}
}

\author{Ying Gao}
\email{gaoying@scut.edu.cn}
\affiliation{%
  \institution{South China University of Technology}
  \city{Guangzhou}
  \state{Guangdong}
  \country{China}
  \postcode{510006}
}

\author{Shanrong Xu}
\email{shanrongxu1@163.com}
\affiliation{%
  \institution{South China University of Technology}
  \city{Guangzhou}
  \state{Guangdong}
  \country{China}
  \postcode{510006}
}


\begin{abstract}
  Nowadays, many classification algorithms have been applied to various industries to help them work out their problems met in real-life scenarios. However, in many binary classification tasks, samples in the minority class only make up a small part of all instances, which leads to the datasets we get usually suffer from high imbalance ratio. Existing models sometimes treat minority classes as noise or ignore them as outliers encountering data skewing. In order to solve this problem, we propose a bagging ensemble learning framework $ASE$ (Anomaly Scoring Based Ensemble Learning). This framework has a scoring system based on anomaly detection algorithms which can guide the resampling strategy by divided samples in the majority class into subspaces. Then specific number of instances will be under-sampled from each subspace to construct subsets by combining with the minority class. And we calculate the weights of base classifiers trained by the subsets according to the classification result of the anomaly detection model and the statistics of the subspaces. Experiments have been conducted which show that our ensemble learning model can dramatically improve the performance of base classifiers and is more efficient than other existing methods under a wide range of imbalance ratio, data scale and data dimension. $ASE$ can be combined with various classifiers and every part of our framework has been proved to be reasonable and necessary.
\end{abstract}

\ccsdesc[500]{Computing methodologies~Bagging}
\ccsdesc[500]{Computing methodologies~Supervised learning by classification}
\ccsdesc[500]{Computing methodologies~Anomaly detection}
\ccsdesc[500]{Computing methodologies~Classification and regression trees}

\keywords{ensemble learning, imbalanced datasets, resampling, anomaly detection, bagging}

\maketitle

\section{Introduction}

Classification is a common research area in machine learning and data mining, and have a wide range of applications in many real-world scenarios\cite{haixiang2017learning}. In these scenarios, our datasets often suffer from high imbalance ratio. Besides data imbalance, data overlapping also accounts for the poor performance of existing models dealing with imbalanced datasets \cite{del2014use}. Confronting with data skewing and data overlapping, classification models are inclined to the majority classes and ignore the minority classes as noise. Models can only get high accuracy on the majority classes but suffer from poor performance on the minority classes. And these kinds of model have no any practical value because people pay more attention to the classification result of the minority class. For example, in credit evaluation, only a few customers are untrustworthy users and the goal is to identify users who may default in the future from all the applicants. If the potential default users are not identified, bad debts will increase and companies will have a direct economic loss, so it is significant to classify the minority class correctly under data skewing situation.

In order to overcome classification problems in data skewing scenarios, resampling methods try to reduce the negative impact of imbalanced data by reducing the high imbalance ratio \cite{chawla2002smote,tomek1976two,smith2014instance,wang2008combination}. But resampling methods have the tendency to ignore some informative instances and discard the original distribution of raw data. And only when the data is not overlapping and can be well clustered, it is feasible to discard or generate some samples. The main idea of cost-sensitive models is reweighting \cite{tang2008svms}. Cost-sensitive models need experts in certain fields to provide prior knowledge like a cost matrix so that models can apply a higher penalty to the minority class to reduce the negative impact of high imbalance ratio, which is not feasible in most situations \cite{krawczyk2013improved}. Some ensemble learning models are likely to be affected by noise and the majority of them only combine resampling methods with multiple base classifiers straightly \cite{chawla2003smoteboost}.

Inspired by the reasons why prevailing methods fail to work out classification tasks under high imbalance ratio and data overlapping, we try to design an innovative ensemble learning framework with a more complicated resample strategy. Instead of using classification error \cite{seiffert2009rusboost,wang2012multiclass,liu2020self} which may lead to error reinforcement, we use anomaly detection to guide the resampling process. Apart from that, we take advantage of the statistics of subsets to weight each base classifier, which makes the final ensemble model can pay more attention to the boundary between the majority class and the minority class.

\begin{figure}[h]
  \centering
  \includegraphics[width=\linewidth]{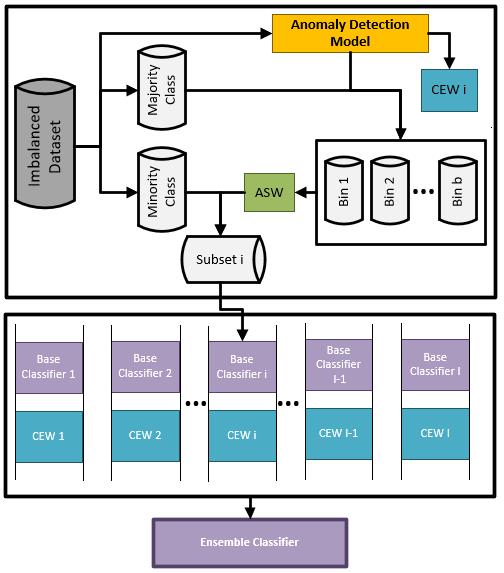}
  \caption{Anomaly Scoring Based Ensemble Learning Model}
  \label{ASE_Frame}
\end{figure}

In this paper, we propose an ensemble learning framework with a scoring system based on anomaly detection algorithms to quantify the anomaly degree of samples. Our framework is a bagging model so it can be trained efficiently with parallel computation. We introduce anomaly detection models to our ensemble Model $ASE$ to quantify the anomaly degree of samples. With higher scores, instances are more likely to be the minority class samples and it is possible for us to detect the overlapping area of the majority class and the minority class. The proposed $ASW$ (Anomaly Scoring Weight) affects the proportion of data splitting and we resample from different subspaces according to $ASW$. $ASW$ depends on the contamination coefficient of anomaly detection models and we use the proposed $CEW$ (Contamination Entropy Weight) to estimate the generalization ability of base classifiers and integrate all weak classifiers into our $ASE$. The pipeline of $ASE$ is shown in Figure~\ref{ASE_Frame}.

In summary, this paper makes these contributions.
\begin{itemize}
\item 1) We introduce a scoring system based on anomaly detection to the resampling step of imbalance classification problems. We use this scoring system to make the proposed ensemble learning framework pay more attention to the minority class and the overlapping area of the two classes so that our framework can achieve higher performance while datasets suffer from severe data skewing.
\item 2) We proposed an ensemble learning framework called $ASE$, which is efficient enough to be applied in a great number of real-life situations to handle classification tasks under high imbalance ratio.
\item 3) We test out our proposed method is much more efficient than other imbalanced classification algorithms on different real-world imbalanced tasks with various base classifiers. With our ensemble framework, we do not need any prior knowledge or pre-defined distance metrics which is inaccessible in most application scenarios.
\end{itemize}


\section{RELATED WORK}
At present, there are three main techniques to deal with imbalanced data, resampling, cost-sensitive learning and ensemble learning \cite{haixiang2017learning}. These methods are widely used in imbalanced classification and Table~\ref{Scenario} shows some application scenarios.

\textbf{Data resampling}:\ Resampling try to solve the problem of data imbalance by reducing datasets' imbalance ratio or even making datasets balanced in the data level. By generating some minority class data (over-sampling) or discarding some majority class data (under-sampling), resampling methods can be divided into three categories, under-sampling, over-sampling and hybrid-sampling. Common resampling methods include RUS \cite{seiffert2009rusboost}, ROS\cite{menardi2014training}, SMOTE\cite{chawla2002smote}, ADASYN\cite{he2008adasyn}, TomekLink\cite{tomek1976two}, OSS\cite{kubat1997addressing}, etc.

Under-sampling tries to reduce imbalance ratio or construct balanced subsets by discarding samples from the majority class. RUS \cite{seiffert2009rusboost} is the oldest under-sampling method which discards samples randomly. Then a serious of under-sampling methods derivated from Nearest Neighbor Criterion including CNN \cite{hart1968condensed}, TomekLink \cite{tomek1976two} and NCR \cite{laurikkala2001improving}. IHT \cite{smith2014instance} removes the instances which have a high possibility to be misclassified by training extra classifiers and OSS \cite{kubat1997addressing} discards samples from the majority class by TomekLink. The idea of evolution and genetic algorithms are also applied to under-sampling like GAUS \cite{ha2016new}. Since under-sampling loses a part of data, it may lead to information loss, which will degrade classifiers' performance. Due to the limitation of the amount of instances, under-sampling is not a wise choice dealing with small scale datasets \cite{haixiang2017learning}.

In order to decrease the imbalance ratio, over-sampling synthesizes new instances based on the minority class\cite{haixiang2017learning}. ROS \cite{menardi2014training} first introduced the idea of over-sampling and SMOTE \cite{chawla2002smote}is the most representative method of over-sampling. Based on SMOTE, variants of SMOTE spring up like Borderline-SMOTE \cite{han2005borderline}, SVM-SMOTE \cite{wang2008combination}. However, over-sampling usually breaks the original data distribution and when data is overlapping and highly imbalanced, this kind of method may not perform well because of overfitting.

Hybrid-sampling is a combination of under-sampling and over-sampling which tries to reduce the negative impact and take advantage of both methods. SMOTE-RSB \cite{ramentol2012smote}and SMOTE-IPF \cite{saez2015smote} are some common hybrid-sampling methods.

\textbf{Cost-sensitive learning}:\ In cost-sensitive learning, a larger penalty coefficient is applied to the minority class so that the classifiers will pay more attention to the classification results of the minority class to reduce the impact of data imbalance. CSSVM \cite{tang2008svms} and CS-LDM \cite{cheng2016cost} are widely used cost-sensitive learning methods. However, cost-sensitive learning requires certain prior knowledge like a cost matrix provided by some experts in specific domains which is not often feasible in real-life situation and models have a certain probability of overfitting \cite{haixiang2017learning}. As the result, compared with resampling and ensemble learning, cost-sensitive learning is a less popular research area and fewer people choose to use it to solve the imbalanced classification problems \cite{krawczyk2013improved}.

\textbf{Ensemble Learning}:\ The main idea of ensemble learning is combining a number of weak classifiers to reconcile the impact of imbalance ratio and noise to a single classifier \cite{haixiang2017learning}. The ensemble method can be divided into two categories, bagging-based ensemble learning models and boosting-based ensemble models. Common ensemble learning methods include UnderBagging \cite{wang2009diversity}, OverBagging \cite{wang2009diversity}, SMOTEBagging \cite{wang2009diversity}, Adaboost\cite{freund1997decision}, SMOTEBoost \cite{chawla2003smoteboost}, RUSBoost \cite{seiffert2009rusboost}, EasyEnsemble \cite{liu2008exploratory} and BalanceCascade \cite{liu2008exploratory}, etc. Ensemble learning improves the generalization ability of the classification models by integrating multiple weak classifiers trained by the data processed by resampling methods, but it is easier to be affected by noise.

Bagging models can be trained parallelly so they are more efficient than boosting models whose base classifiers need to be trained based on the information given by the last base classifier. Some experts \cite{fernandez2013analysing} have verified that bagging methods' performance doesn't fall behind boosting methods while costing less time in training so these methods are widely applied in practical applications. Bagging algorithms include OverBagging \cite{wang2009diversity}, UnderBagging \cite{wang2009diversity}and SMOTEBagging \cite{wang2009diversity}, etc. BPSO \cite{li2022binary} introduces an under-sampling method named Binary PSO to select instances used to train base classifiers. HUE \cite{ng2020hashing} uses a hashing method to divide samples in the majority class into subspaces. SubFeat \cite{haque2021subfeat} divides features into overlapping and non-overlapping spaces and uses subspaces to train individual classifier. EPX \cite{hsu2021epx} scores all rare class subjects by clustering feature space and exploiting the richness of features. DELAK \cite{yang2021ensemble} introduces a concept called the distance-based dynamic ensemble which can combine the output of base classifiers for new testing samples dynamically.

Boosting models take advantage of the heuristic information like the classification results during the iterative process and can be only trained in sequential process. Adaboost first took the idea of boosting into practice and more boosting algorithms are designed afterwards like SMOTEBoost \cite{chawla2003smoteboost}, RUSBoost \cite{seiffert2009rusboost}, AdaboostNC \cite{wang2012multiclass}, BSIA \cite{zikeba2015boosted}, etc. Both EasyEnsemble and BalanceCascade \cite{liu2008exploratory} use Adaboost as their base classifiers and BalanceCascade adjusts the threshold depending on the false positive rate. ERFADASYN \cite{balaram2022prediction} proposes a new feature selection called Butterfly optimization to solve the class imbalance problem. In Enslia \cite{jing2022ensemble}, the “excellent and diverse” principle helps to guide the training process of base classifiers. SPE \cite{liu2020self} and MESA \cite{liu2020mesa} are boosting models trained by subsets undersampled according to hardness distributions \cite{li2019gradient}.

\section{Proposed Method}


\subsection{Symbol definition}

In this paper, we define the majority class as the negative class and $N$ is the set of all negative samples, the minority class as the positive class and $P$ is the set of all positive samples. The definitions are shown in Equation~\ref{def_pos} and Equation~\ref{def_neg}.

\begin{equation}
\label{def_pos}
P = \{(x,y)|y=1\}
\end{equation}

\begin{equation}
\label{def_neg}
N=\{(x,y)|y=0\}
\end{equation}

IR (Imbalance Ratio) is used to quantify the level of data skewing in a datase which is show in Equation~\ref{equ_ir}.
\begin{equation}
\label{equ_ir}
IR = \frac{the\ number\ of\ majority\ class\ samples}{the\ number\ of\ minority\ class\ samples} = \frac{|N|}{|P|}
\end{equation}

And we take Confusion Matrix which is shown in Table~\ref{Confusion_Matrix} to evaluate the anomaly detection models' performance on the training set and the final performance of our ensemble framework.

\begin{table}
\caption{Confusion Matrix}
\label{Confusion_Matrix}
\begin{tabular}{lcc}
\toprule
 & Predict Positive & Predict Negative \\
\midrule
Label Positive & TP & FN\\ 
Label Negative & FP & TN\\	
\bottomrule
\end{tabular}
\end{table}

$x_j$ is any sample in the dataset and $A_i$ is the $i$-th anomaly detection model. $AS_{i,j} = A_i(x_j)$ denotes the anomaly score of $x_j$ and $B_{i,l}$ represents the $l$-th bin in the $i$-th base classifier. $ASW_{i,l}$ is the weight of $B_{i,l}$ and $CEW_{i}$ is the weight of the $i$-th weak classifier. $C$ is our Contamination function which is relevant to the percentage of outliers of the anomaly detection model.

\subsection{Anomaly Scoring System}


		


Since under-sampling is easy to suffer from discarding some informative instances and disturb underlying distributions, some try to extract some hidden information from the raw data so that they applied Nearest Neighbor to guide resampling like CNN \cite{hart1968condensed} and NCR \cite{laurikkala2001improving}. Nearest Neighbor is often used in clustering and anomaly detection, so we try to apply other advanced algorithms in Anomaly Detection like Isolation Forest \cite{liu2008isolation}, SVDD \cite{ruff2018deep} etc. Anomaly detection models will score the examples and examples whose scores surpass the threshold will be classified as the outliers. In $ASE$, we used these scores to guide our resampling strategy. Some methods introduce a concept called Hardness Distribution \cite{li2019gradient,liu2020self,liu2020mesa} and they use some common functions like Absolute Error, Squared Error or Cross Entropy as their hardness function which try to represent the hardness to classify examples correctly. And then they split all positive samples into different bins and take the mean of all examples in a specified bin as the weight of this bin.

Inspired by the idea of instances’ weight and splitting data into bins, now we introduce our own concept called $ASW$. Our $ASW$ needs one hyper-parameter $k$ which is the number of bins. We split all training samples into $k$ bins by the anomaly score $AS$ in Equation~\ref{equ_as} given by the anomaly detection model $A$ and each bin indicates a certain anomaly level as shown in Equation~\ref{splitting}.

\begin{equation}
\label{equ_as}
AS_{i,j} = A_i(x_j)
\end{equation}
\begin{equation}
\label{splitting}
B_{i,l} = \{(x_j,y_j)|\frac{l-1}{k}\leq AS_{i,j} < \frac{l}{k}\}
\end{equation}

Since our datasets are under high imbalance ratio, a part of the minority class will be regarded as the outliers, which have a greater probability to receive higher anomaly scores from the anomaly detection model than most samples in the majority class. However, what makes the imbalanced classification intractable is that the majority class usually overlaps with the minority class \cite{del2014use} and there is no a specific boundary between the two classes. The overlapping data will seriously affect the performance of the base classifier so we need to pay more attention to these samples with a weighting function shown in Equation~\ref{equ_ASW}.

\begin{equation}
\label{equ_ASW}
ASW_{i,l} = \frac{ log\frac{1}{|B_{i,l}|}}{\sum_{l=1}^{k}log(\frac{1}{|B_{i,l}|})} 
\end{equation}
	
\begin{equation}
\label{equ_d}
d_{i,l} = \frac{1}{|\frac{l}{k}-\frac{1}{2k}-(1-c_i)|}
\end{equation}

$ASW_{i,l}$ represents the weight of the $l$-th bin according to the $i$-th anomaly detection model and the number of instances needed to be resampled from the majority class of the $l$-th bin is $n_{i,l}$. $c_i$ is the contamination coefficient used to train the $i$-th base classifier which is equal to the percentage of samples which the anomaly detection model classifies as the outliers and in Equation~\ref{equ_d}, $d_{i,l}$ implies the distance between the $l$-th bin and the boundary of the outliers.

\begin{equation}
n_{i,l} = |N|\cdot ASW_{i,l} \cdot \frac{d_{i,l}}{\sum_{l=1}^k d_{i,l}}
\end{equation}

Then we combine the resampled instances from different bins with all the samples from the minority class to construct a subset which has a lower imbalance ratio than the original dataset to train a base classifier.

\begin{figure}[h]
  \centering
  \includegraphics[width=\linewidth]{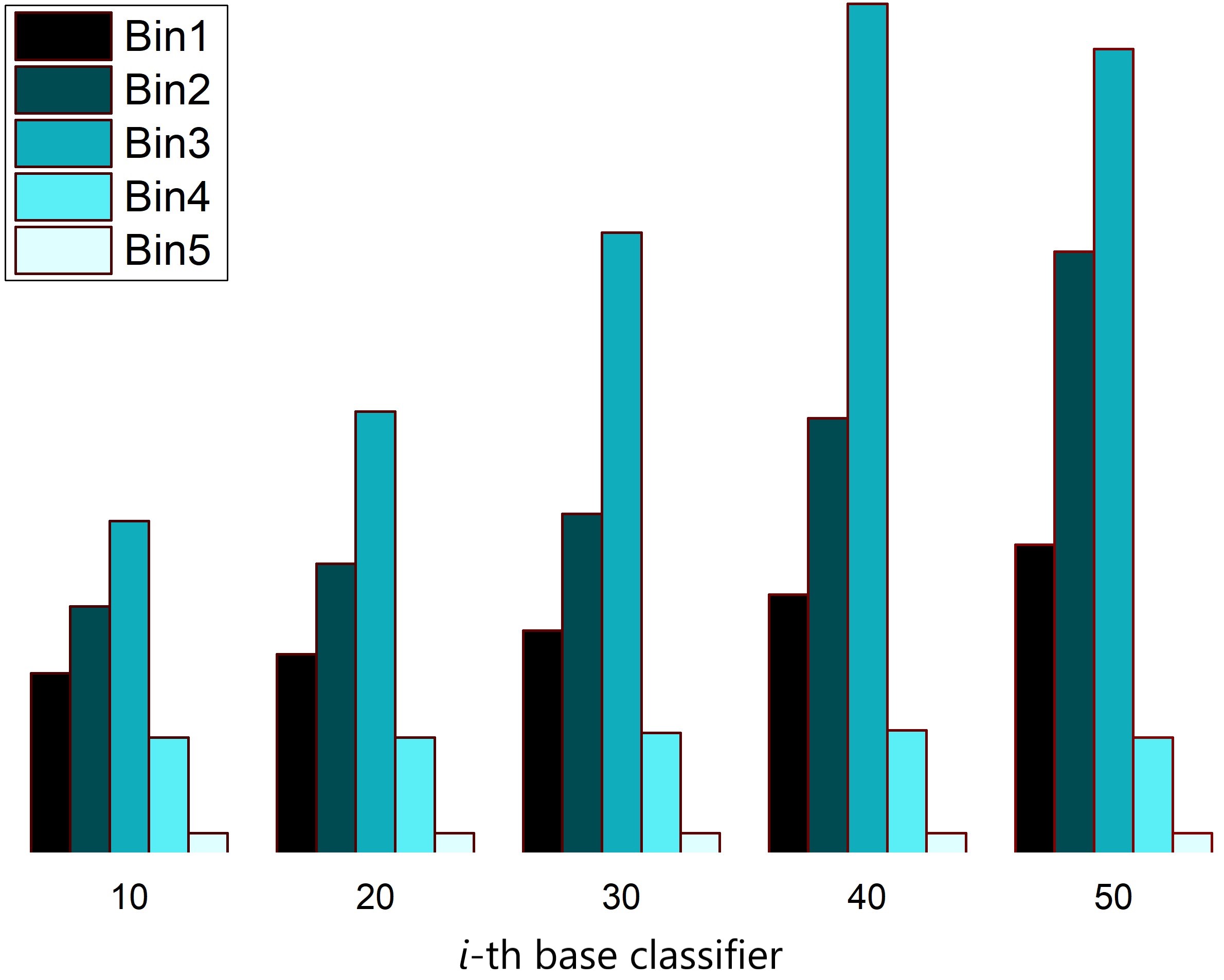}
  \caption{An example to show how $ASW$ affects the process of splitting. We combine 50 weak classifiers to train an $ASE$ model using dataset $Wine$}
  \label{ASE}
\end{figure}


Figure~\ref{ASE} shows how $ASW$ works to decide the resampling proportion in $ASE$. We train 50 weak classifiers with the contamination coefficient increasing from 0.05 to 0.40 to construct our ensemble learning model on dataset $Wine$ which includes about 5,000 instances and the imbalance ratio is about 26:1. The number of samples in each bin is displayed and we choose the distributions of data in 5 rounds of training to be shown in Figure~\ref{ASE}.

\subsection{Contamination Entropy Weight Ensemble}

When constructing weak classifiers into an ensemble model in many mainstream methods, they will not combine the weak classifiers directly but get the weighted average of all weak classifiers. The weight functions are usually relative to the classification results of the base classifiers generated in the previous training process \cite{zhang2022intelligent}. In RUSBoost \cite{seiffert2009rusboost}, the pseudo-loss will be calculated to update the weight parameters which will have a great impact on the resampling method in the next iteration and will directly affect the weight of the base classifier trained in this iteration. AdaboostNC \cite{wang2012multiclass} tries to access the disagreement degree of the classification with penalty strength and calculates the weight of base classifiers by error and penalty. Some factors derivated from the confusion matrix like the weight of FP and the quotient of TP and FP instances are applied to update the model in each iteration. In this paper, we propose our weight function \emph{CEW}.

While training each base classifier, we change the contamination coefficient of the anomaly detection model, which is relative to the amount of contaminated instances of the dataset. Since the proportion of the outliers is changed, the Recall on the training set will be changed. With higher Recall, our anomaly detection model has a higher performance on the minority class and the minority instances are more possible to be splitted into the bin with higher $ASW$. In this case, more instances in the minority class will be splitted into bins with higher $ASW$ so the number of instances resampled from these bins will increase. In result, the $n_{i,l}$ where $l$ is more closed to $k$ will increase, that is to say, the base classifiers will focus more on more informative instances in the majority class like the overlapping area between the majority class and the minority class.

The contamination percentage will directly affect the Recall value, but the main idea to attach an anomaly detection model to our framework is not only to detect all the outliers and classify the minority class correctly the first time, but also to balance the proportion of the number of the majority class, the minority class and the overlapping data in the constructed subsets. In order to handle the data overlapping problems, we need our model to pay more attention to the overlapping area in the original dataset. In $ASE$, we combine the idea of information entropy and confusion matrix to weigh each base classifier and the proposed weight function $CEW$ is shown in Equation~\ref{equ_CEW}.

\begin{equation}
\label{equ_CEW}
CEW_{i} = C_i \cdot E_i
\end{equation}

$CEW$ can be divided into two parts. As shown in Equation~\ref{equ_C}, $C$ is a contamination function relative to Recall, which reveals the anomaly detection model’s ability to detect the minority class exactly. In Equation~\ref{equ_E}, $E$ is a function derivated from information entropy that can quantitatively assess the anomaly model generalization ability on the overlapping area between majority class and minority class.

\begin{equation}
\label{equ_C}
\begin{aligned}
C &= log(\frac{|P|}{FN})\\
  &= log(\frac{1}{1-Recall})
\end{aligned}
\end{equation}

\begin{equation}
\label{equ_E}
\begin{aligned}
		E &= - (\sum _{i=1}^{k}p_i log(p_i))\\
		  &= - \frac{TP}{TP+FP} \cdot log(\frac{TP}{TP+FP}) \\
		  & \ \ \ \ - \frac{FP}{TP+FP} \cdot log(\frac{FP}{TP+FP})
\end{aligned}
\end{equation}

The core idea of information entropy \cite{shannon1948mathematical} is to quantify the value of information. When there is not enough information concealed in data or something is highly likely to happen, we say that the information entropy in this case is low. When the majority class or the minority class consist of the most part of the outliers detected by the anomaly detection model, we can regard the result of the anomaly detection model can not give enough information for us to find the overlapping area between the majority and minority class and this situation has a low information entropy. On the contrary, when the number of the majority class in examples predicted as positive (FP) is closed to the number of the minority class in examples predicted as positive (TP), the examples in the overlapping area are nearly balanced and we will receive a higher information entropy. Since the overlapping area is more balanced, the weak classifier trained in this iteration will have a better generalization ability of the overlapping data so we should assign a greater weight to it.

\subsection{ASE (Anomaly Scoring Ensemble Learning)}

After discussing our model’s modules in detail, now we will give the general framework of $ASE$. Firstly, we need an anomaly detection model like Isolation Forest \cite{liu2008isolation}, SVDD \cite{ruff2018deep} or Auto Encoder \cite{zhou2017anomaly} to assign anomaly scores $AS$ to the training data. Then we need to split the majority class into $k$ bins according to anomaly scores and we will calculate $ASW$ for each bin. With $ASW$ we can down-sample a specific number of instances from each bin and combine these instances with minority class to build a subset that has adjusted the proportion among the majority class, the minority class and the overlapping area. In this way, we make our classification model pay appropriate attention to different bins based on their contribution to datasets' chaos. Base classifiers trained by these subsets can suffer less from high imbalance ratio and disruption of the raw data distributions by resampling and have a better generalization ability. In each training process of a new base classifier, the proportions of contaminated data are set differently which will affect the resampling strategy directly and we need to quantify each base classifier's generalization ability so we introduce our weight function $CEW$. $CEW$ consists of two parts which are relative to the ability of the anomaly detection model to find the minority class $C$ and the ability to discover the overlapping area between the two classes $E$. At last, we combine all the base classifiers into a strong classifier which can adapt to various degrees of imbalance ratio, data scale and feature dimension with our $CEW$ function. The framework of $ASE$ is shown in Figure~\ref{ASE_Frame} and the details of $ASE$ is shown in Algorithm~\ref{algASE}.

\begin{algorithm}[tb]
   \caption{ASE (Anomaly Scoring Based Ensemble Learning)}
   \label{algASE}
\begin{algorithmic}[1]
   \STATE {\bfseries Input:} minority class $P$, majority class $N$, base classifier $M$, anomaly detection model $A$, number of bins $k$, number of base classifiers $b$
   \FOR{$i=1$ {\bfseries to} $b$}
   \STATE Train anomaly detection model $A_i$ with contamination coefficient $c_i$ and normalize $AS_i$
   \STATE Split training set into $k$ bins w.r.t. $AS_{i,}$
   \STATE Compute $ASW$ for each bin and under-sample specific number of examples from the majority class in each bin
   \STATE Construct subset $s_i$ by combining the resampled data and all the data in the minority class
   \STATE Train base classifier $M_i$ with subset $s_i$
   \STATE Compute $CEW_i$ for $M_i$ 
   \ENDFOR
   \STATE Ensemble base classifiers $ASE = \sum_{i=1}^{b}M_i \cdot CEW_i$
\end{algorithmic}
\end{algorithm}

\section{Experiment}

\subsection{Parameter Setting and Evaluation Criterion}

In all our experiments, we randomly choose 80\% of samples from the majority class and the minority class respectively as the training set and the samples left are in the testing set.


If it is not specified, the training set is splitted into 5 bins and the depth of all tree models is set as 10, which are used in \cite{liu2020self, liu2020mesa} and can help us to compare our model with others. Apart from that, all ensemble learning models contain 50 base classifiers, which is the default value in Python package imbalanced-learn \cite{JMLR:v18:16-365} and other parameters are the default values in imbalanced-learn \cite{JMLR:v18:16-365} and scikit-learn \cite{scikit-learn}. In other words, our model's two main hyper-parameters $k$ is set as 5 and $b$ is set as 50 by default.

We use some common criteria in imbalance learning area based on the confusion matrix. If the classification model considers all the testing samples as the majority class, the model will still get a high accuracy score because of the high imbalance ratio. Precision and recall are usually used to evaluate models’ performance on the minority class. So we take Accuracy, Precision, Recall, AUC and F1 to evaluate our model performance.


	


\subsection{Datasets}
There are many application scenarios for imbalanced classification and various algorithms are widely used in different industries, like finance, manufacturing, scientific research and medicine. A great number of areas suffer from data imbalance in real-life situations such as risk prediction, financial assessment, loan default, credit card fraud, network intrusion detection, fraud detection, product quality inspection and disease diagnosis. Therefore, we selected several datasets which are representative in each scenario to evaluate our proposed framework.

Table~\ref{Datasets_Detail} contains the statistics of all datasets used in this paper.

\begin{table}[]
\caption{Datasets Detail}
\label{Datasets_Detail}
\begin{tabular}{lcccc}
\toprule
 Dataset & Repository & Instance & Feature & {IR} \\
\midrule
	Credit Fault & Kaggle & 284,807 & 31 & 579:1 \\
	Credit & Kaggle & 150,000 & 11 & 14:1 \\
	KDD2004 & KDD Cup & 145,751 & 74 & 111:1 \\
	\midrule
	Letter & UCI & 20,000 & 16 & 26:1 \\
	Mammography & UCI & 11,183 & 6 & 42:1 \\
    \midrule
	ISOLET & UCI & 7,797 & 617 & 12:1 \\
	Wine & UCI & 4,898 & 11 & 26:1 \\
	Ozone Level & UCI & 2,536 & 72 & 34:1 \\
\bottomrule
\end{tabular}
\end{table}

\begin{itemize}

\item \textbf{Credit Fault:}\ The dataset (Credit Card Fraud Detection) \cite{dal2017credit} contains transactions made by credit cards in September 2013 by European cardholders. Each entity has 30 features, which are the result of a PCA transformation and this dataset contains 492 frauds out of 284,807 transactions. It is significant for credit card companies to recognize fraudulent credit card transactions under a high imbalance ratio about 579:1.

\item \textbf{Credit:}\ Give Me Some Credit is a dataset about credit scoring used in Kaggle featured prediction competition. We randomly select 150,000 instances with the imbalance ratio of 13.96:1 for our experiment. Each instance has 11 features about credit card users' personal information.

\item \textbf{KDD2004:}\ This dataset is provided by KDD Cup 2004, focusing on predicting which proteins are homologous to a native sequence. This dataset has 145,751 instances with 74 features and the imbalance ratio 111:1.

\item \textbf{Letter:}\ There are a large number of black-and-white rectangular pixel displays as 26 capital letters in UCI dataset Letter Recognition. This dataset has 20,000 instances with 16 features and the imbalance ratio 26:1.

\item \textbf{Mammography:}\ Mammography is the most effective method for breast cancer screening available today. Each instance in Mammography (Mammographic Mass Data Set) only contains 6 features and it has only 260 malignant instances out of 11,183 records.

\item \textbf{ISOLET:}\ ISOLET (Isolated Letter Speech Recognition) is a real-world dataset generated by 150 subjects who spoke the name of each letter of the alphabet. ISOLET is a high-dimension dataset with 617 attributes, containing 7,797 instances and 600 instances among which are the minority class.

\item \textbf{Wine:}\ Wine Quality Data Set is a UCI \cite{asuncion2007uci} dataset, including two datasets related to red and white variants of the Portuguese "Vinho Verde" wine. Since there are more normal wines than excellent or poor ones, this dataset has a imbalance ratio about 26:1 and 4,898 instances with 11 features.

\item \textbf{Ozone Level:}\ UCI dataset Ozone Level contains 2,536 instances with 72 dimensions, which were collected from 1998 to 2004 at the Houston, Galveston and Brazoria area. This imbalanced dataset with imbalance ratio 34:1 quantifies two ground ozone levels.

\end{itemize}

\subsection{Experiment Design and Result}

\subsubsection{Experiments on Different Models}

\begin{table*}[]
\caption{Experiments on 6 datasets}
\label{real_world_result}
\begin{tabular}{{llccccccccccccc}}
\toprule
  &           & \multicolumn{13}{c}{Algorithm}  \\
\toprule
Dataset & Metric & DT & SMOTE & RF & GBDT &
\begin{tabular}{@{}c@{}}Ada- \\ boost\end{tabular} & SPE & Balance& Easy & 
\begin{tabular}{@{}c@{}}RUS- \\ Boost\end{tabular} & 
\begin{tabular}{@{}c@{}}SMOTE- \\ Boost\end{tabular} &
\begin{tabular}{@{}c@{}}Under\end{tabular} &
\begin{tabular}{@{}c@{}}Over\end{tabular} & \textbf{ASE} \\
\midrule

\multirow{5}{*}{\begin{tabular}{@{}c@{}}Credit\\ Fault\end{tabular}} & Acc  & \textbf{0.999} & 0.991 & \textbf{0.999} & \textbf{0.999} & \textbf{0.999} & \textbf{0.999} & \textbf{0.999} & 0.977          & 0.942    & 0.988      & 0.981          & \textbf{0.999} & \emph{\textbf{0.999}} \\  & Precision & 0.899          & 0.146 & 0.941 & 0.796          & 0.833          & 0.796          & 0.650          & 0.065          & 0.028    & 0.120      & 0.077          &  \emph{\textbf{0.959}}          & 0.840          \\& Recall    & 0.756          & 0.832 & 0.784          & 0.760          & 0.695          & 0.839          & 0.852          & \textbf{0.887} & 0.876    & 0.875      & 0.885          & 0.738          & 0.873          \\
                              & AUC       & 0.878          & 0.912 & 0.892          & 0.880          & 0.848          & 0.919          & 0.925          & 0.932          & 0.909    & 0.932      & 0.933          & 0.869          & \emph{\textbf{0.936}} \\
                              & F1        & 0.820          & 0.247 & \textbf{0.856} & 0.777          & 0.757          & 0.816          & 0.737          & 0.121          & 0.054    & 0.210      & 0.141          & 0.833          & \emph{\textbf{0.856}}          \\
\hline
\multirow{5}{*}{ISOLET}       & Acc  & 0.939          & 0.933 & 0.965          & 0.974          & 0.971          & 0.977          & 0.970          & 0.928          & 0.891    & 0.957      & 0.951          & 0.963          & \emph{\textbf{0.984}} \\
                              & Precision & 0.579          & 0.525 & 0.942          & \textbf{0.949} & 0.817          & 0.784          & 0.722          & 0.507          & 0.378    & 0.668      & 0.609          & 0.811          & 0.822          \\
                              & Recall    & 0.618          & 0.828 & 0.555          & 0.685          & 0.774          & 0.952          & 0.954          & 0.941          & 0.725    & 0.824      & 0.936          & 0.653          & \emph{\textbf{0.990}} \\
                              & AUC       & 0.791          & 0.884 & 0.776          & 0.841          & 0.880          & 0.966          & 0.962          & 0.934          & 0.814    & 0.896      & 0.944          & 0.820          & \emph{\textbf{0.987}} \\
                              & F1        & 0.596          & 0.642 & 0.697          & 0.795          & 0.794          & 0.859          & 0.821          & 0.658          & 0.494    & 0.737      & 0.737          & 0.721          & \emph{\textbf{0.898}} \\
\hline
\multirow{5}{*}{\begin{tabular}{@{}c@{}}KDD \\ 2004\end{tabular}}      & Acc  & \textbf{0.997} & 0.973 & \textbf{0.997} & 0.996          & \textbf{0.997} & \textbf{0.997} & 0.995          & 0.965          & 0.948    & 0.974      & 0.979          & \textbf{0.997} & \emph{\textbf{0.997}} \\
                              & Precision & 0.906          & 0.226 & \textbf{0.981} & 0.799          & 0.859          & 0.851          & 0.670          & 0.186          & 0.133    & 0.233      & 0.275          & 0.959          & 0.810          \\
                              & Recall    & 0.689          & 0.878 & 0.659          & 0.746          & 0.726          & 0.846          & 0.876          & \textbf{0.927} & 0.879    & 0.904      & 0.922          & 0.708          & 0.914          \\
                              & AUC       & 0.844          & 0.926 & 0.830          & 0.872          & 0.863          & 0.923          & 0.936          & 0.946          & 0.914    & 0.939      & 0.950          & 0.854          & \emph{\textbf{0.956}} \\
                              & F1        & 0.783          & 0.360 & 0.788          & 0.771          & 0.786          & 0.848          & 0.758          & 0.310          & 0.230    & 0.370      & 0.423          & 0.814          & \emph{\textbf{0.859}} \\
\hline
\multirow{5}{*}{Mam}  & Acc  & 0.983          & 0.964 & 0.987          & 0.986          & 0.985          & 0.974          & 0.677          & 0.903          & 0.875    & 0.932      & 0.946          & 0.986          & \emph{\textbf{0.990}} \\
                              & Precision & 0.666          & 0.363 & \textbf{0.894}          & 0.785          & 0.754          & 0.473          & 0.059          & 0.175          & 0.139    & 0.234      & 0.283          & 0.841 & 0.744          \\
                              & Recall    & 0.551          & 0.728 & 0.506          & 0.548          & 0.498          & 0.810          & \textbf{0.863} & 0.848          & 0.764    & 0.831      & 0.852          & 0.483          & 0.853          \\
                              & AUC       & 0.772          & 0.848 & 0.752          & 0.772          & 0.747          & 0.894          & 0.768          & 0.876          & 0.821    & 0.883      & 0.900          & 0.740          & \emph{\textbf{0.923}} \\
                              & F1        & 0.597          & 0.483 & 0.645          & 0.642          & 0.599          & 0.595          & 0.111          & 0.289          & 0.233    & 0.364      & 0.425          & 0.613          & \emph{\textbf{0.793}} \\
\hline
\multirow{5}{*}{Letter}       & Acc  & 0.994          & 0.991 & 0.996          & \textbf{0.997} & 0.993          & \textbf{0.997} & 0.996          & 0.944          & 0.962    & 0.986      & 0.984          & \textbf{0.997} & \emph{\textbf{0.997}}          \\
                              & Precision & 0.928          & 0.840 & \textbf{0.997} & 0.981          & 0.919          & 0.968          & 0.912          & 0.388          & 0.504    & 0.733      & 0.698          & 0.980          & 0.957          \\
                              & Recall    & 0.917          & 0.935 & 0.887          & 0.938          & 0.887          & 0.955          & 0.970          & 0.975          & 0.926    & 0.954      & \textbf{0.977} & 0.929          & 0.962          \\
                              & AUC       & 0.957          & 0.964 & 0.943          & 0.969          & 0.942          & 0.977          & \textbf{0.980} & 0.959          & 0.945    & 0.971      & \textbf{0.980}          & 0.964          & \emph{\textbf{0.980}} \\
                              & F1        & 0.922          & 0.884 & 0.938          & 0.959 & 0.903          & \emph{\textbf{0.960}}          & 0.940          & 0.555          & 0.643    & 0.829      & 0.814          & 0.953          & \emph{\textbf{0.960}} \\
                              
\hline
\multirow{5}{*}{Ozone}       
& Acc  & 0.949        & 0.91         & 0.97         & 0.955 & 0.968    & 0.869 & 0.842          & 0.862        & 0.847    & 0.935      & 0.874        & 0.971       & \emph{\textbf{0.982}} \\

& Precision & 0.192        & 0.157        & 0.1          & 0.235 & 0.383    & 0.163 & 0.132          & 0.15         & 0.09     & 0.241      & 0.156        & 0.592       & \emph{\textbf{0.632}} \\

& Recall   &  0.213        & 0.484        & 0.007        & 0.217 & 0.218    & 0.811 & 0.77           & 0.778        & 0.428    & 0.557      & 0.739        & 0.085       & \emph{\textbf{0.953}} \\

& AUC       & 0.592        & 0.704        & 0.503        & 0.598 & 0.605    & 0.841 & 0.807          & 0.822        & 0.644    & 0.752      & 0.808        & 0.542       & \emph{\textbf{0.968}} \\

& F1        & 0.191        & 0.234        & 0.013        & 0.216 & 0.271    & 0.268 & 0.224          & 0.249        & 0.147    & 0.327      & 0.256        & 0.14        & \emph{\textbf{0.754}} \\

\bottomrule
\end{tabular}
\end{table*}

We first use Isolation Forest as the anomaly detection model and Decision Tree as the base classifier in our experiment. In order to test out the performance of $ASE$, we use 6 datasets, Credit Fault, ISOLET, KDD2004, Mammography, Letter and Ozone Level to compare $ASE$ with 12 existing methods, including resampling methods and ensemble learning models which are widely used in imbalanced datasets. In detail, the models we select include Decision Tree, SMOTE, Random Forest, GBDT, Adaboost, UnderBagging, OverBagging, EasyEnsemble, BalanceCascade, SMOTEBoost, RUSBoost and SPE.

The results are shown in Table~\ref{real_world_result} and $ASE$ gets the highest scores of $Accuracy$, $AUC$ and $F1$ among all 13 compared methods in 6 representative datasets in various scenarios, where data usually suffers from high imbalance ratio. Compared with the second-best F1 score in Ozone Level and Mammography, $ASE$ achieves 181\% and 23\% performance gain. 

The experiment results show that $ASE$ is well-designed and has an exceeding ability to work out the data skewing problem than existing methods. The fluctuation of imbalance ratio, the scale of datasets and the changes in application scenarios demonstrate $ASE$ not only has high performance but also can be used in various situations.

\subsubsection{Experiments on Different Base Classifiers and Anomaly Detection Algorithms}
The above experiment uses Decision Tree as the base classifier and Isolation Forest as the anomaly detection model. In order to illustrate the excellent design philosophy of $ASE$, we change the base classifiers and use various anomaly detection algorithms in $ASE$.

To compare $ASE$ performance while using different base classifiers, We then use Decision Tree, Logistic Regression, SVM, KNN and MLP in datasets KDD2004, Credit and Wine to compare $ASE$ with SMOTE, SPE, EasyEnsemble and BalanceCascade and use $F1$ as evaluation criteria. As shown in Table~\ref{classifier_cmp}, $ASE$ gets all the highest $F1$ which demonstrates that $ASE$ is not only limited to a specific classifier, this ensemble learning framework has a good generalization ability which can be applied to various classifiers.

\begin{table*}[]
\caption{Performance of ASE with Different Base Classifiers}
\label{classifier_cmp}
\begin{tabular}{lccccccc}
\toprule
 Dataset    &  Base Classifer    & None  & SMOTE & SPE   & \begin{tabular}{@{}c@{}}Balance- \\ Cascade\end{tabular}  & \begin{tabular}{@{}c@{}}Easy- \\ Ensemble\end{tabular}   & \textbf{ASE}   \\
\midrule
\multirow{5}{*}{KDD2004} & DT            & 0.798 & 0.345 & 0.841 & 0.743   & 0.304 & \textit{\textbf{0.866}} \\
                         & LR            & 0.778 & 0.227 & 0.763 & 0.660   & 0.213 & \textit{\textbf{0.778}} \\
                         & SVM           & 0.643 & 0.255 & 0.740 & 0.615   & 0.121 & \textit{\textbf{0.747}} \\
                         & KNN           & 0.560 & 0.239 & 0.148 & 0.178   & 0.098 & \textit{\textbf{0.599}} \\
                         & MLP           & 0.828 & 0.764 & 0.811 & 0.334   & 0.705 & \textit{\textbf{0.828}} \\
\hline
\multirow{3}{*}{Credit Fault}  & DT            & 0.288 & 0.340 & 0.357 & 0.179   & 0.323 & \textit{\textbf{0.363}} \\
                         & KNN           & 0.041 & 0.158 & 0.160 & 0.133   & 0.174 & \textit{\textbf{0.245}} \\
                         & MLP           & 0.126 & 0.300 & 0.140 & 0.272   & 0.360 & \textit{\textbf{0.380}} \\
\hline
\multirow{3}{*}{Wine}    & DT & 0.277 & 0.208 & 0.254 & 0.256   & 0.229 & \textit{\textbf{0.549}} \\
                         & KNN           & 0.067 & 0.170 & 0.135 & 0.082   & 0.150 & \textit{\textbf{0.376}} \\
                         & MLP           & 0.096 & 0.203 & 0.198 & 0.189   & 0.111 & \textit{\textbf{0.236}} \\
\bottomrule
\end{tabular}
\end{table*}

\begin{table*}[]
\caption{Performance of ASE with Different Anomaly Detection Model on Dataset $Wine$}
\label{AD_cmp}
\begin{tabular}{lccccccccc}
\toprule
          & DT  & UnderBagging & BalanceCascade & SPE & Iforest-ASE & OCSVM-ASE & KNN-ASE & AE-ASE & ROD-ASE \\
\midrule
Accuracy  & 0.855 & 0.872 & 0.846          & 0.835 & 0.958       & 0.928          & 0.974          & 0.957  & \textbf{0.971} \\
Precision & 0.130 & 0.182 & 0.155          & 0.151 & 0.448       & 0.338          & \textbf{0.649} & 0.492  & 0.592          \\
Recall    & 0.531 & 0.745 & 0.737          & 0.789 & 0.703       & \textbf{0.855} & 0.657          & 0.701  & 0.605          \\
AUC       & 0.699 & 0.811 & 0.793          & 0.813 & 0.835       & \textbf{0.893} & 0.822          & 0.834  & 0.794          \\
F1        & 0.208 & 0.292 & 0.256          & 0.254 & 0.545       & 0.472          & \textbf{0.651} & 0.576  & 0.595   \\
\bottomrule
\end{tabular}
\end{table*}

Apart from changing the base classifiers, the experiment which combines different anomaly detection models with $ASE$ has been conducted. We use some representative and efficient anomaly detection models including Isolation Forest, KNN, OCSVM, Auto Encoder and ROD as the anomaly scoring part of $ASE$.

\begin{itemize}
\item \textbf{Isolation Forest} \cite{liu2008isolation} randomly selects a value between the minimum and maximum values to split the data into partitions since anomalous data are few and different.
\item \textbf{OCSVM} \cite{manevitz2001one} is an unsupervised outlier detection which can estimate the support of a high-dimensional distribution and is based on the premise that outliers will cluster in a dense region in the original dataset.
\item \textbf{KNN}  \cite{angiulli2002fast} is a popular algorithm widely used in classification, anomaly detection which assumes that outliers usually stay away from the cluster of similar instances. 

\item \textbf{Auto Encoder} \cite{zhou2017anomaly} can be used to detect anomaly by encoding and compressing the data into the lower dimensions and then decode to reconstruct the data.
\item \textbf{ROD} \cite{almardeny2020novel} is a parameter-free algorithm which uses 3D-vectors to represent the raw data and uses Rodrigues rotation formula for scoring the data to find the outliers.
    
\end{itemize}

Decision Tree is selected as the base classifier trained on dataset $Wine$ and the results are shown in Table~\ref{AD_cmp}. We compare $ASE$ with SMOTE, EasyEnsemble, BalanceCascade, UnderBagging, OverBagging and SPE but we only show the compared models with the highest performance including UnderBagging, OverBagging and SPE because of the layout.

\subsubsection{Experiments the Sensitivity to Hyper-parameters}
Since the number of base classifiers is the key determinant of the performance of ensemble learning models, we compare the training process of $ASE$ with other ensemble learning models, including EasyEnsemble, BalanceCascade and $SPE$. Besides, we also change the hyper-parameter $k$ to split the training set into different number of bin, for example, $k$ is set as 3 in $ASE3$. We choose dataset $Wine$ and $ISOLET$ and Decision Tree as the base classifier.

\begin{figure}[ht]
\begin{center}

	\subfigure[Wine AUC]{
		\label{level.sub.1}
		\includegraphics[width=0.45\linewidth]{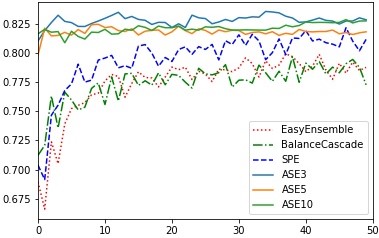}}
\quad 
	\subfigure[Wine F1]{
		\label{level.sub.2}
		\includegraphics[width=0.45\linewidth]{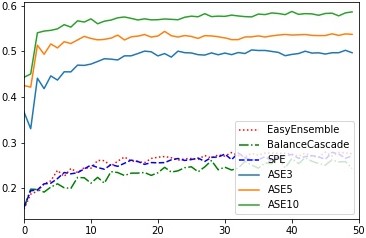}}
		
	\subfigure[ISOLET AUC]{
		\label{level.sub.3}
		\includegraphics[width=0.45\linewidth]{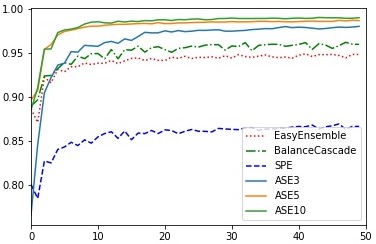}}
\quad 
	\subfigure[ISOLET F1]{
		\label{level.sub.4}
		\includegraphics[width=0.45\linewidth]{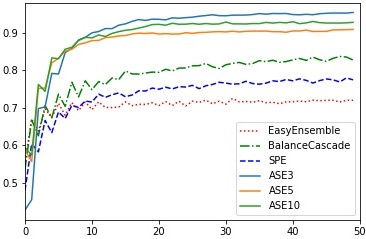}}

\caption{Generalized Performance of $ASE$}
\label{generalization}
\end{center}
\vskip -0.2in
\end{figure}

\begin{table*}[]
\caption{Ablation Experiment}
\label{Ablation Experiment}
\begin{tabular}{lcccc}
\toprule
Dataset      & ASE without CEW & ASE without ASW & ASE without both & ASE   \\
\midrule
Credit Fault & 0.836           & 0.829           & 0.816            & \textit{\textbf{0.856}} \\
Mammography  & 0.788           & 0.668           & 0.590            & \textit{\textbf{0.791}} \\
Ozone        & 0.742           & 0.553           & 0.549            & \textit{\textbf{0.747}} \\
Wine         & 0.515           & 0.490           & 0.413            & \textit{\textbf{0.520}} \\
\bottomrule
\end{tabular}
\end{table*}

Figure~\ref{generalization} shows the changing process of the $ROC$ and $F1$ when training ensemble learning models, $ASE$ nearly performs best during all the training process. $ASE$ performs well with limited base classifiers and converges faster than mainstream ensemble learning models. At the same time, it is robust to different selection of $k$.

\subsubsection{Ablation Experiment}
In order to verify that our $ASE$ is well designed and is not redundant, we conduct an ablation experiment to show that each module in $ASE$ like $CEW$ and $ASW$ is reasonable and necessary. We use Decision Tree as the base classifier and combine 20 base classifiers into the final ensemble model. In the ablation experiment, we assign equal weights to each base classifier to test out whether $CEW$ boosts the performance of $ASE$ or not. $ASW$ is discarded and we use a new strategy to resample while ensuring the $CEW$ is working at the same time. We don't use the splitting strategy in Equation~\ref{splitting} and we split the training set into bins by the quantiles of the output anomaly scores. As a result, the number of examples in each bin is the same so we down-sample randomly in each bin to construct a new subset. Besides, we also set up a control group without both $ASW$ and $CEW$. We use F1 as the criterion of the ablation experiment.

The result of the ablation experiment is shown in Table~\ref{Ablation Experiment}. The removal of either $ASW$ or $CEW$ from our framework has a negative impact on $ASE$ and it seems that $ASW$ has a greater impact on $ASE$ than $CEW$ since it directly changes the resampling strategy, which makes the proposed framework a success far beyond the existing models. Since $CEW$ is determined by the proportion of the samples that the anomaly model classifies as the positive class, which contains some informative information like the information entropy and the Recall value of the anomaly detection model, $CEW$ can help to improve the performance of the ensemble model.

\section{Conclusions}
High imbalance ratio, data overlapping, large scale and high feature dimension are common intractable problems in many real-life classification scenarios. Existing models like machine learning methods or ensemble models applied in these scenarios suffer from poor performance because of flaws embedded in their design. In this paper, we illustrate how data overlapping affects the performance of classification algorithms and we propose an innovative ensemble learning framework, Anomaly Scoring Based Ensemble Learning to work out the classification problems in data skewing real-life scenarios. Since our model can evaluate the overlapping area of the majority class and the minority class efficiently, the ensemble model has a better generalization ability on highly imbalanced datasets with high-dimension features. Experiments on several datasets including finance, medicine, manufacturing industry and meteorology have been conducted and the performance and applicability of our model have been tested out. Besides, we have designed an ablation experiment to approve each module in our framework can boost the ensemble model's performance and our model is logical and not redundant. We believe that our ensemble learning model can be applied to various real-life scenarios.

\bibliographystyle{ACM-Reference-Format}
\bibliography{ASE}


\begin{thebibliography}{48}


\ifx \showCODEN    \undefined \def \showCODEN     #1{\unskip}     \fi
\ifx \showDOI      \undefined \def \showDOI       #1{#1}\fi
\ifx \showISBNx    \undefined \def \showISBNx     #1{\unskip}     \fi
\ifx \showISBNxiii \undefined \def \showISBNxiii  #1{\unskip}     \fi
\ifx \showISSN     \undefined \def \showISSN      #1{\unskip}     \fi
\ifx \showLCCN     \undefined \def \showLCCN      #1{\unskip}     \fi
\ifx \shownote     \undefined \def \shownote      #1{#1}          \fi
\ifx \showarticletitle \undefined \def \showarticletitle #1{#1}   \fi
\ifx \showURL      \undefined \def \showURL       {\relax}        \fi
\providecommand\bibfield[2]{#2}
\providecommand\bibinfo[2]{#2}
\providecommand\natexlab[1]{#1}
\providecommand\showeprint[2][]{arXiv:#2}

\bibitem[Almardeny et~al\mbox{.}(2020)]%
        {almardeny2020novel}
\bibfield{author}{\bibinfo{person}{Yahya Almardeny},
  \bibinfo{person}{Noureddine Boujnah}, {and} \bibinfo{person}{Frances
  Cleary}.} \bibinfo{year}{2020}\natexlab{}.
\newblock \showarticletitle{A novel outlier detection method for multivariate
  data}.
\newblock \bibinfo{journal}{\emph{IEEE Transactions on Knowledge and Data
  Engineering}} (\bibinfo{year}{2020}).
\newblock


\bibitem[Angiulli and Pizzuti(2002)]%
        {angiulli2002fast}
\bibfield{author}{\bibinfo{person}{Fabrizio Angiulli} {and}
  \bibinfo{person}{Clara Pizzuti}.} \bibinfo{year}{2002}\natexlab{}.
\newblock \showarticletitle{Fast outlier detection in high dimensional spaces}.
  In \bibinfo{booktitle}{\emph{European conference on principles of data mining
  and knowledge discovery}}. Springer, \bibinfo{pages}{15--27}.
\newblock


\bibitem[Asuncion and Newman(2007)]%
        {asuncion2007uci}
\bibfield{author}{\bibinfo{person}{Arthur Asuncion} {and}
  \bibinfo{person}{David Newman}.} \bibinfo{year}{2007}\natexlab{}.
\newblock \bibinfo{title}{UCI machine learning repository}.
\newblock
\newblock


\bibitem[Balaram and Vasundra(2022)]%
        {balaram2022prediction}
\bibfield{author}{\bibinfo{person}{A Balaram} {and} \bibinfo{person}{S
  Vasundra}.} \bibinfo{year}{2022}\natexlab{}.
\newblock \showarticletitle{Prediction of software fault-prone classes using
  ensemble random forest with adaptive synthetic sampling algorithm}.
\newblock \bibinfo{journal}{\emph{Automated Software Engineering}}
  \bibinfo{volume}{29}, \bibinfo{number}{1} (\bibinfo{year}{2022}),
  \bibinfo{pages}{1--21}.
\newblock


\bibitem[Chawla et~al\mbox{.}(2002)]%
        {chawla2002smote}
\bibfield{author}{\bibinfo{person}{Nitesh~V Chawla}, \bibinfo{person}{Kevin~W
  Bowyer}, \bibinfo{person}{Lawrence~O Hall}, {and} \bibinfo{person}{W~Philip
  Kegelmeyer}.} \bibinfo{year}{2002}\natexlab{}.
\newblock \showarticletitle{SMOTE: synthetic minority over-sampling technique}.
\newblock \bibinfo{journal}{\emph{Journal of artificial intelligence research}}
   \bibinfo{volume}{16} (\bibinfo{year}{2002}), \bibinfo{pages}{321--357}.
\newblock


\bibitem[Chawla et~al\mbox{.}(2003)]%
        {chawla2003smoteboost}
\bibfield{author}{\bibinfo{person}{Nitesh~V Chawla},
  \bibinfo{person}{Aleksandar Lazarevic}, \bibinfo{person}{Lawrence~O Hall},
  {and} \bibinfo{person}{Kevin~W Bowyer}.} \bibinfo{year}{2003}\natexlab{}.
\newblock \showarticletitle{SMOTEBoost: Improving prediction of the minority
  class in boosting}. In \bibinfo{booktitle}{\emph{European conference on
  principles of data mining and knowledge discovery}}. Springer,
  \bibinfo{pages}{107--119}.
\newblock


\bibitem[Cheng et~al\mbox{.}(2016)]%
        {cheng2016cost}
\bibfield{author}{\bibinfo{person}{Fanyong Cheng}, \bibinfo{person}{Jing
  Zhang}, {and} \bibinfo{person}{Cuihong Wen}.}
  \bibinfo{year}{2016}\natexlab{}.
\newblock \showarticletitle{Cost-sensitive large margin distribution machine
  for classification of imbalanced data}.
\newblock \bibinfo{journal}{\emph{Pattern Recognition Letters}}
  \bibinfo{volume}{80} (\bibinfo{year}{2016}), \bibinfo{pages}{107--112}.
\newblock


\bibitem[Dal~Pozzolo et~al\mbox{.}(2017)]%
        {dal2017credit}
\bibfield{author}{\bibinfo{person}{Andrea Dal~Pozzolo},
  \bibinfo{person}{Giacomo Boracchi}, \bibinfo{person}{Olivier Caelen},
  \bibinfo{person}{Cesare Alippi}, {and} \bibinfo{person}{Gianluca Bontempi}.}
  \bibinfo{year}{2017}\natexlab{}.
\newblock \showarticletitle{Credit card fraud detection: a realistic modeling
  and a novel learning strategy}.
\newblock \bibinfo{journal}{\emph{IEEE transactions on neural networks and
  learning systems}} \bibinfo{volume}{29}, \bibinfo{number}{8}
  (\bibinfo{year}{2017}), \bibinfo{pages}{3784--3797}.
\newblock


\bibitem[Del~R{\'\i}o et~al\mbox{.}(2014)]%
        {del2014use}
\bibfield{author}{\bibinfo{person}{Sara Del~R{\'\i}o},
  \bibinfo{person}{Victoria L{\'o}pez}, \bibinfo{person}{Jos{\'e}~Manuel
  Ben{\'\i}tez}, {and} \bibinfo{person}{Francisco Herrera}.}
  \bibinfo{year}{2014}\natexlab{}.
\newblock \showarticletitle{On the use of mapreduce for imbalanced big data
  using random forest}.
\newblock \bibinfo{journal}{\emph{Information Sciences}}  \bibinfo{volume}{285}
  (\bibinfo{year}{2014}), \bibinfo{pages}{112--137}.
\newblock


\bibitem[Fern{\'a}ndez et~al\mbox{.}(2013)]%
        {fernandez2013analysing}
\bibfield{author}{\bibinfo{person}{Alberto Fern{\'a}ndez},
  \bibinfo{person}{Victoria L{\'o}Pez}, \bibinfo{person}{Mikel Galar},
  \bibinfo{person}{Mar{\'\i}A~Jos{\'e} Del~Jesus}, {and}
  \bibinfo{person}{Francisco Herrera}.} \bibinfo{year}{2013}\natexlab{}.
\newblock \showarticletitle{Analysing the classification of imbalanced
  data-sets with multiple classes: Binarization techniques and ad-hoc
  approaches}.
\newblock \bibinfo{journal}{\emph{Knowledge-based systems}}
  \bibinfo{volume}{42} (\bibinfo{year}{2013}), \bibinfo{pages}{97--110}.
\newblock


\bibitem[Freund and Schapire(1997)]%
        {freund1997decision}
\bibfield{author}{\bibinfo{person}{Yoav Freund} {and} \bibinfo{person}{Robert~E
  Schapire}.} \bibinfo{year}{1997}\natexlab{}.
\newblock \showarticletitle{A decision-theoretic generalization of on-line
  learning and an application to boosting}.
\newblock \bibinfo{journal}{\emph{Journal of computer and system sciences}}
  \bibinfo{volume}{55}, \bibinfo{number}{1} (\bibinfo{year}{1997}),
  \bibinfo{pages}{119--139}.
\newblock


\bibitem[Ha and Lee(2016)]%
        {ha2016new}
\bibfield{author}{\bibinfo{person}{Jihyun Ha} {and} \bibinfo{person}{Jong-Seok
  Lee}.} \bibinfo{year}{2016}\natexlab{}.
\newblock \showarticletitle{A new under-sampling method using genetic algorithm
  for imbalanced data classification}. In \bibinfo{booktitle}{\emph{Proceedings
  of the 10th International Conference on Ubiquitous Information Management and
  Communication}}. \bibinfo{pages}{1--6}.
\newblock


\bibitem[Haixiang et~al\mbox{.}(2017)]%
        {haixiang2017learning}
\bibfield{author}{\bibinfo{person}{Guo Haixiang}, \bibinfo{person}{Li Yijing},
  \bibinfo{person}{Jennifer Shang}, \bibinfo{person}{Gu Mingyun},
  \bibinfo{person}{Huang Yuanyue}, {and} \bibinfo{person}{Gong Bing}.}
  \bibinfo{year}{2017}\natexlab{}.
\newblock \showarticletitle{Learning from class-imbalanced data: Review of
  methods and applications}.
\newblock \bibinfo{journal}{\emph{Expert systems with applications}}
  \bibinfo{volume}{73} (\bibinfo{year}{2017}), \bibinfo{pages}{220--239}.
\newblock


\bibitem[Han et~al\mbox{.}(2005)]%
        {han2005borderline}
\bibfield{author}{\bibinfo{person}{Hui Han}, \bibinfo{person}{Wen-Yuan Wang},
  {and} \bibinfo{person}{Bing-Huan Mao}.} \bibinfo{year}{2005}\natexlab{}.
\newblock \showarticletitle{Borderline-SMOTE: a new over-sampling method in
  imbalanced data sets learning}. In \bibinfo{booktitle}{\emph{International
  conference on intelligent computing}}. Springer, \bibinfo{pages}{878--887}.
\newblock


\bibitem[Haque et~al\mbox{.}(2021)]%
        {haque2021subfeat}
\bibfield{author}{\bibinfo{person}{HM~Fazlul Haque}, \bibinfo{person}{Muhammod
  Rafsanjani}, \bibinfo{person}{Fariha Arifin}, \bibinfo{person}{Sheikh
  Adilina}, {and} \bibinfo{person}{Swakkhar Shatabda}.}
  \bibinfo{year}{2021}\natexlab{}.
\newblock \showarticletitle{Subfeat: Feature subspacing ensemble classifier for
  function prediction of dna, rna and protein sequences}.
\newblock \bibinfo{journal}{\emph{Computational Biology and Chemistry}}
  \bibinfo{volume}{92} (\bibinfo{year}{2021}), \bibinfo{pages}{107489}.
\newblock


\bibitem[Hart(1968)]%
        {hart1968condensed}
\bibfield{author}{\bibinfo{person}{Peter Hart}.}
  \bibinfo{year}{1968}\natexlab{}.
\newblock \showarticletitle{The condensed nearest neighbor rule (corresp.)}.
\newblock \bibinfo{journal}{\emph{IEEE transactions on information theory}}
  \bibinfo{volume}{14}, \bibinfo{number}{3} (\bibinfo{year}{1968}),
  \bibinfo{pages}{515--516}.
\newblock


\bibitem[He et~al\mbox{.}(2008)]%
        {he2008adasyn}
\bibfield{author}{\bibinfo{person}{Haibo He}, \bibinfo{person}{Yang Bai},
  \bibinfo{person}{Edwardo~A Garcia}, {and} \bibinfo{person}{Shutao Li}.}
  \bibinfo{year}{2008}\natexlab{}.
\newblock \showarticletitle{ADASYN: Adaptive synthetic sampling approach for
  imbalanced learning}. In \bibinfo{booktitle}{\emph{2008 IEEE international
  joint conference on neural networks (IEEE world congress on computational
  intelligence)}}. IEEE, \bibinfo{pages}{1322--1328}.
\newblock


\bibitem[Hsu et~al\mbox{.}(2021)]%
        {hsu2021epx}
\bibfield{author}{\bibinfo{person}{Grace~G Hsu}, \bibinfo{person}{Jabed~H
  Tomal}, {and} \bibinfo{person}{William~J Welch}.}
  \bibinfo{year}{2021}\natexlab{}.
\newblock \showarticletitle{EPX: An R package for the ensemble of subsets of
  variables for highly unbalanced binary classification}.
\newblock \bibinfo{journal}{\emph{Computers in Biology and Medicine}}
  \bibinfo{volume}{136} (\bibinfo{year}{2021}), \bibinfo{pages}{104760}.
\newblock


\bibitem[Jing et~al\mbox{.}(2022)]%
        {jing2022ensemble}
\bibfield{author}{\bibinfo{person}{Chao Jing}, \bibinfo{person}{Yun Wu}, {and}
  \bibinfo{person}{Chaoyuan Cui}.} \bibinfo{year}{2022}\natexlab{}.
\newblock \showarticletitle{Ensemble dynamic behavior detection method for
  adversarial malware}.
\newblock \bibinfo{journal}{\emph{Future Generation Computer Systems}}
  \bibinfo{volume}{130} (\bibinfo{year}{2022}), \bibinfo{pages}{193--206}.
\newblock


\bibitem[Krawczyk and Schaefer(2013)]%
        {krawczyk2013improved}
\bibfield{author}{\bibinfo{person}{Bartosz Krawczyk} {and}
  \bibinfo{person}{Gerald Schaefer}.} \bibinfo{year}{2013}\natexlab{}.
\newblock \showarticletitle{An improved ensemble approach for imbalanced
  classification problems}. In \bibinfo{booktitle}{\emph{2013 IEEE 8th
  international symposium on applied computational intelligence and informatics
  (SACI)}}. IEEE, \bibinfo{pages}{423--426}.
\newblock


\bibitem[Kubat et~al\mbox{.}(1997)]%
        {kubat1997addressing}
\bibfield{author}{\bibinfo{person}{Miroslav Kubat}, \bibinfo{person}{Stan
  Matwin}, {et~al\mbox{.}}} \bibinfo{year}{1997}\natexlab{}.
\newblock \showarticletitle{Addressing the curse of imbalanced training sets:
  one-sided selection}. In \bibinfo{booktitle}{\emph{Icml}},
  Vol.~\bibinfo{volume}{97}. Citeseer, \bibinfo{pages}{179}.
\newblock


\bibitem[Laurikkala(2001)]%
        {laurikkala2001improving}
\bibfield{author}{\bibinfo{person}{Jorma Laurikkala}.}
  \bibinfo{year}{2001}\natexlab{}.
\newblock \showarticletitle{Improving identification of difficult small classes
  by balancing class distribution}. In \bibinfo{booktitle}{\emph{Conference on
  artificial intelligence in medicine in Europe}}. Springer,
  \bibinfo{pages}{63--66}.
\newblock


\bibitem[Lema{{\^i}}tre et~al\mbox{.}(2017)]%
        {JMLR:v18:16-365}
\bibfield{author}{\bibinfo{person}{Guillaume Lema{{\^i}}tre},
  \bibinfo{person}{Fernando Nogueira}, {and} \bibinfo{person}{Christos~K.
  Aridas}.} \bibinfo{year}{2017}\natexlab{}.
\newblock \showarticletitle{Imbalanced-learn: A Python Toolbox to Tackle the
  Curse of Imbalanced Datasets in Machine Learning}.
\newblock \bibinfo{journal}{\emph{Journal of Machine Learning Research}}
  \bibinfo{volume}{18}, \bibinfo{number}{17} (\bibinfo{year}{2017}),
  \bibinfo{pages}{1--5}.
\newblock
\urldef\tempurl%
\url{http://jmlr.org/papers/v18/16-365.html}
\showURL{%
\tempurl}


\bibitem[Li et~al\mbox{.}(2019)]%
        {li2019gradient}
\bibfield{author}{\bibinfo{person}{Buyu Li}, \bibinfo{person}{Yu Liu}, {and}
  \bibinfo{person}{Xiaogang Wang}.} \bibinfo{year}{2019}\natexlab{}.
\newblock \showarticletitle{Gradient harmonized single-stage detector}. In
  \bibinfo{booktitle}{\emph{Proceedings of the AAAI conference on artificial
  intelligence}}, Vol.~\bibinfo{volume}{33}. \bibinfo{pages}{8577--8584}.
\newblock


\bibitem[Li et~al\mbox{.}(2022)]%
        {li2022binary}
\bibfield{author}{\bibinfo{person}{Jinyan Li}, \bibinfo{person}{Yaoyang Wu},
  \bibinfo{person}{Simon Fong}, \bibinfo{person}{Antonio~J
  Tall{\'o}n-Ballesteros}, \bibinfo{person}{Xin-she Yang},
  \bibinfo{person}{Sabah Mohammed}, {and} \bibinfo{person}{Feng Wu}.}
  \bibinfo{year}{2022}\natexlab{}.
\newblock \showarticletitle{A binary PSO-based ensemble under-sampling model
  for rebalancing imbalanced training data}.
\newblock \bibinfo{journal}{\emph{The Journal of Supercomputing}}
  \bibinfo{volume}{78}, \bibinfo{number}{5} (\bibinfo{year}{2022}),
  \bibinfo{pages}{7428--7463}.
\newblock


\bibitem[Liu et~al\mbox{.}(2008a)]%
        {liu2008isolation}
\bibfield{author}{\bibinfo{person}{Fei~Tony Liu}, \bibinfo{person}{Kai~Ming
  Ting}, {and} \bibinfo{person}{Zhi-Hua Zhou}.}
  \bibinfo{year}{2008}\natexlab{a}.
\newblock \showarticletitle{Isolation forest}. In
  \bibinfo{booktitle}{\emph{2008 eighth ieee international conference on data
  mining}}. IEEE, \bibinfo{pages}{413--422}.
\newblock


\bibitem[Liu et~al\mbox{.}(2008b)]%
        {liu2008exploratory}
\bibfield{author}{\bibinfo{person}{Xu-Ying Liu}, \bibinfo{person}{Jianxin Wu},
  {and} \bibinfo{person}{Zhi-Hua Zhou}.} \bibinfo{year}{2008}\natexlab{b}.
\newblock \showarticletitle{Exploratory undersampling for class-imbalance
  learning}.
\newblock \bibinfo{journal}{\emph{IEEE Transactions on Systems, Man, and
  Cybernetics, Part B (Cybernetics)}} \bibinfo{volume}{39}, \bibinfo{number}{2}
  (\bibinfo{year}{2008}), \bibinfo{pages}{539--550}.
\newblock


\bibitem[Liu et~al\mbox{.}(2020a)]%
        {liu2020self}
\bibfield{author}{\bibinfo{person}{Zhining Liu}, \bibinfo{person}{Wei Cao},
  \bibinfo{person}{Zhifeng Gao}, \bibinfo{person}{Jiang Bian},
  \bibinfo{person}{Hechang Chen}, \bibinfo{person}{Yi Chang}, {and}
  \bibinfo{person}{Tie-Yan Liu}.} \bibinfo{year}{2020}\natexlab{a}.
\newblock \showarticletitle{Self-paced ensemble for highly imbalanced massive
  data classification}. In \bibinfo{booktitle}{\emph{2020 IEEE 36th
  international conference on data engineering (ICDE)}}. IEEE,
  \bibinfo{pages}{841--852}.
\newblock


\bibitem[Liu et~al\mbox{.}(2020b)]%
        {liu2020mesa}
\bibfield{author}{\bibinfo{person}{Zhining Liu}, \bibinfo{person}{Pengfei Wei},
  \bibinfo{person}{Jing Jiang}, \bibinfo{person}{Wei Cao},
  \bibinfo{person}{Jiang Bian}, {and} \bibinfo{person}{Yi Chang}.}
  \bibinfo{year}{2020}\natexlab{b}.
\newblock \showarticletitle{MESA: boost ensemble imbalanced learning with
  meta-sampler}.
\newblock \bibinfo{journal}{\emph{Advances in Neural Information Processing
  Systems}}  \bibinfo{volume}{33} (\bibinfo{year}{2020}),
  \bibinfo{pages}{14463--14474}.
\newblock


\bibitem[Manevitz and Yousef(2001)]%
        {manevitz2001one}
\bibfield{author}{\bibinfo{person}{Larry~M Manevitz} {and}
  \bibinfo{person}{Malik Yousef}.} \bibinfo{year}{2001}\natexlab{}.
\newblock \showarticletitle{One-class SVMs for document classification}.
\newblock \bibinfo{journal}{\emph{Journal of machine Learning research}}
  \bibinfo{volume}{2}, \bibinfo{number}{Dec} (\bibinfo{year}{2001}),
  \bibinfo{pages}{139--154}.
\newblock


\bibitem[Menardi and Torelli(2014)]%
        {menardi2014training}
\bibfield{author}{\bibinfo{person}{Giovanna Menardi} {and}
  \bibinfo{person}{Nicola Torelli}.} \bibinfo{year}{2014}\natexlab{}.
\newblock \showarticletitle{Training and assessing classification rules with
  imbalanced data}.
\newblock \bibinfo{journal}{\emph{Data mining and knowledge discovery}}
  \bibinfo{volume}{28}, \bibinfo{number}{1} (\bibinfo{year}{2014}),
  \bibinfo{pages}{92--122}.
\newblock


\bibitem[Ng et~al\mbox{.}(2020)]%
        {ng2020hashing}
\bibfield{author}{\bibinfo{person}{Wing~WY Ng}, \bibinfo{person}{Shichao Xu},
  \bibinfo{person}{Jianjun Zhang}, \bibinfo{person}{Xing Tian},
  \bibinfo{person}{Tongwen Rong}, {and} \bibinfo{person}{Sam Kwong}.}
  \bibinfo{year}{2020}\natexlab{}.
\newblock \showarticletitle{Hashing-based undersampling ensemble for imbalanced
  pattern classification problems}.
\newblock \bibinfo{journal}{\emph{IEEE Transactions on Cybernetics}}
  (\bibinfo{year}{2020}).
\newblock


\bibitem[Pedregosa et~al\mbox{.}(2011)]%
        {scikit-learn}
\bibfield{author}{\bibinfo{person}{F. Pedregosa}, \bibinfo{person}{G.
  Varoquaux}, \bibinfo{person}{A. Gramfort}, \bibinfo{person}{V. Michel},
  \bibinfo{person}{B. Thirion}, \bibinfo{person}{O. Grisel},
  \bibinfo{person}{M. Blondel}, \bibinfo{person}{P. Prettenhofer},
  \bibinfo{person}{R. Weiss}, \bibinfo{person}{V. Dubourg}, \bibinfo{person}{J.
  Vanderplas}, \bibinfo{person}{A. Passos}, \bibinfo{person}{D. Cournapeau},
  \bibinfo{person}{M. Brucher}, \bibinfo{person}{M. Perrot}, {and}
  \bibinfo{person}{E. Duchesnay}.} \bibinfo{year}{2011}\natexlab{}.
\newblock \showarticletitle{Scikit-learn: Machine Learning in {P}ython}.
\newblock \bibinfo{journal}{\emph{Journal of Machine Learning Research}}
  \bibinfo{volume}{12} (\bibinfo{year}{2011}), \bibinfo{pages}{2825--2830}.
\newblock


\bibitem[Ramentol et~al\mbox{.}(2012)]%
        {ramentol2012smote}
\bibfield{author}{\bibinfo{person}{Enislay Ramentol},
  \bibinfo{person}{Yail{\'e} Caballero}, \bibinfo{person}{Rafael Bello}, {and}
  \bibinfo{person}{Francisco Herrera}.} \bibinfo{year}{2012}\natexlab{}.
\newblock \showarticletitle{SMOTE-RSB*: a hybrid preprocessing approach based
  on oversampling and undersampling for high imbalanced data-sets using SMOTE
  and rough sets theory}.
\newblock \bibinfo{journal}{\emph{Knowledge and information systems}}
  \bibinfo{volume}{33}, \bibinfo{number}{2} (\bibinfo{year}{2012}),
  \bibinfo{pages}{245--265}.
\newblock


\bibitem[Ruff et~al\mbox{.}(2018)]%
        {ruff2018deep}
\bibfield{author}{\bibinfo{person}{Lukas Ruff}, \bibinfo{person}{Robert
  Vandermeulen}, \bibinfo{person}{Nico Goernitz}, \bibinfo{person}{Lucas
  Deecke}, \bibinfo{person}{Shoaib~Ahmed Siddiqui}, \bibinfo{person}{Alexander
  Binder}, \bibinfo{person}{Emmanuel M{\"u}ller}, {and} \bibinfo{person}{Marius
  Kloft}.} \bibinfo{year}{2018}\natexlab{}.
\newblock \showarticletitle{Deep one-class classification}. In
  \bibinfo{booktitle}{\emph{International conference on machine learning}}.
  PMLR, \bibinfo{pages}{4393--4402}.
\newblock


\bibitem[S{\'a}ez et~al\mbox{.}(2015)]%
        {saez2015smote}
\bibfield{author}{\bibinfo{person}{Jos{\'e}~A S{\'a}ez},
  \bibinfo{person}{Juli{\'a}n Luengo}, \bibinfo{person}{Jerzy Stefanowski},
  {and} \bibinfo{person}{Francisco Herrera}.} \bibinfo{year}{2015}\natexlab{}.
\newblock \showarticletitle{SMOTE--IPF: Addressing the noisy and borderline
  examples problem in imbalanced classification by a re-sampling method with
  filtering}.
\newblock \bibinfo{journal}{\emph{Information Sciences}}  \bibinfo{volume}{291}
  (\bibinfo{year}{2015}), \bibinfo{pages}{184--203}.
\newblock


\bibitem[Seiffert et~al\mbox{.}(2009)]%
        {seiffert2009rusboost}
\bibfield{author}{\bibinfo{person}{Chris Seiffert}, \bibinfo{person}{Taghi~M
  Khoshgoftaar}, \bibinfo{person}{Jason Van~Hulse}, {and} \bibinfo{person}{Amri
  Napolitano}.} \bibinfo{year}{2009}\natexlab{}.
\newblock \showarticletitle{RUSBoost: A hybrid approach to alleviating class
  imbalance}.
\newblock \bibinfo{journal}{\emph{IEEE Transactions on Systems, Man, and
  Cybernetics-Part A: Systems and Humans}} \bibinfo{volume}{40},
  \bibinfo{number}{1} (\bibinfo{year}{2009}), \bibinfo{pages}{185--197}.
\newblock


\bibitem[Shannon(1948)]%
        {shannon1948mathematical}
\bibfield{author}{\bibinfo{person}{Claude~Elwood Shannon}.}
  \bibinfo{year}{1948}\natexlab{}.
\newblock \showarticletitle{A mathematical theory of communication}.
\newblock \bibinfo{journal}{\emph{The Bell system technical journal}}
  \bibinfo{volume}{27}, \bibinfo{number}{3} (\bibinfo{year}{1948}),
  \bibinfo{pages}{379--423}.
\newblock


\bibitem[Smith et~al\mbox{.}(2014)]%
        {smith2014instance}
\bibfield{author}{\bibinfo{person}{Michael~R Smith}, \bibinfo{person}{Tony
  Martinez}, {and} \bibinfo{person}{Christophe Giraud-Carrier}.}
  \bibinfo{year}{2014}\natexlab{}.
\newblock \showarticletitle{An instance level analysis of data complexity}.
\newblock \bibinfo{journal}{\emph{Machine learning}} \bibinfo{volume}{95},
  \bibinfo{number}{2} (\bibinfo{year}{2014}), \bibinfo{pages}{225--256}.
\newblock


\bibitem[Tang et~al\mbox{.}(2008)]%
        {tang2008svms}
\bibfield{author}{\bibinfo{person}{Yuchun Tang}, \bibinfo{person}{Yan-Qing
  Zhang}, \bibinfo{person}{Nitesh~V Chawla}, {and} \bibinfo{person}{Sven
  Krasser}.} \bibinfo{year}{2008}\natexlab{}.
\newblock \showarticletitle{SVMs modeling for highly imbalanced
  classification}.
\newblock \bibinfo{journal}{\emph{IEEE Transactions on Systems, Man, and
  Cybernetics, Part B (Cybernetics)}} \bibinfo{volume}{39}, \bibinfo{number}{1}
  (\bibinfo{year}{2008}), \bibinfo{pages}{281--288}.
\newblock


\bibitem[Tomek(1976)]%
        {tomek1976two}
\bibfield{author}{\bibinfo{person}{Ivan Tomek}.}
  \bibinfo{year}{1976}\natexlab{}.
\newblock \showarticletitle{Two modifications of CNN}.
\newblock \bibinfo{journal}{\emph{IEEE Trans. Systems, Man and Cybernetics}}
  \bibinfo{volume}{6} (\bibinfo{year}{1976}), \bibinfo{pages}{769--772}.
\newblock


\bibitem[Wang(2008)]%
        {wang2008combination}
\bibfield{author}{\bibinfo{person}{He-Yong Wang}.}
  \bibinfo{year}{2008}\natexlab{}.
\newblock \showarticletitle{Combination approach of SMOTE and biased-SVM for
  imbalanced datasets}. In \bibinfo{booktitle}{\emph{2008 IEEE International
  Joint Conference on Neural Networks (IEEE World Congress on Computational
  Intelligence)}}. IEEE, \bibinfo{pages}{228--231}.
\newblock


\bibitem[Wang and Yao(2009)]%
        {wang2009diversity}
\bibfield{author}{\bibinfo{person}{Shuo Wang} {and} \bibinfo{person}{Xin Yao}.}
  \bibinfo{year}{2009}\natexlab{}.
\newblock \showarticletitle{Diversity analysis on imbalanced data sets by using
  ensemble models}. In \bibinfo{booktitle}{\emph{2009 IEEE symposium on
  computational intelligence and data mining}}. IEEE,
  \bibinfo{pages}{324--331}.
\newblock


\bibitem[Wang and Yao(2012)]%
        {wang2012multiclass}
\bibfield{author}{\bibinfo{person}{Shuo Wang} {and} \bibinfo{person}{Xin Yao}.}
  \bibinfo{year}{2012}\natexlab{}.
\newblock \showarticletitle{Multiclass imbalance problems: Analysis and
  potential solutions}.
\newblock \bibinfo{journal}{\emph{IEEE Transactions on Systems, Man, and
  Cybernetics, Part B (Cybernetics)}} \bibinfo{volume}{42}, \bibinfo{number}{4}
  (\bibinfo{year}{2012}), \bibinfo{pages}{1119--1130}.
\newblock


\bibitem[Yang et~al\mbox{.}(2021)]%
        {yang2021ensemble}
\bibfield{author}{\bibinfo{person}{Ping Yang}, \bibinfo{person}{Dan Wang},
  \bibinfo{person}{Wen-Bing Zhao}, \bibinfo{person}{Li-Hua Fu},
  \bibinfo{person}{Jin-Lian Du}, {and} \bibinfo{person}{Hang Su}.}
  \bibinfo{year}{2021}\natexlab{}.
\newblock \showarticletitle{Ensemble of kernel extreme learning machine based
  random forest classifiers for automatic heartbeat classification}.
\newblock \bibinfo{journal}{\emph{Biomedical Signal Processing and Control}}
  \bibinfo{volume}{63} (\bibinfo{year}{2021}), \bibinfo{pages}{102138}.
\newblock


\bibitem[Zhang et~al\mbox{.}(2022)]%
        {zhang2022intelligent}
\bibfield{author}{\bibinfo{person}{Tianci Zhang}, \bibinfo{person}{Jinglong
  Chen}, \bibinfo{person}{Fudong Li}, \bibinfo{person}{Kaiyu Zhang},
  \bibinfo{person}{Haixin Lv}, \bibinfo{person}{Shuilong He}, {and}
  \bibinfo{person}{Enyong Xu}.} \bibinfo{year}{2022}\natexlab{}.
\newblock \showarticletitle{Intelligent fault diagnosis of machines with small
  \& imbalanced data: A state-of-the-art review and possible extensions}.
\newblock \bibinfo{journal}{\emph{ISA transactions}}  \bibinfo{volume}{119}
  (\bibinfo{year}{2022}), \bibinfo{pages}{152--171}.
\newblock


\bibitem[Zhou and Paffenroth(2017)]%
        {zhou2017anomaly}
\bibfield{author}{\bibinfo{person}{Chong Zhou} {and} \bibinfo{person}{Randy~C
  Paffenroth}.} \bibinfo{year}{2017}\natexlab{}.
\newblock \showarticletitle{Anomaly detection with robust deep autoencoders}.
  In \bibinfo{booktitle}{\emph{Proceedings of the 23rd ACM SIGKDD international
  conference on knowledge discovery and data mining}}.
  \bibinfo{pages}{665--674}.
\newblock


\bibitem[Zi{\k{e}}ba and Tomczak(2015)]%
        {zikeba2015boosted}
\bibfield{author}{\bibinfo{person}{Maciej Zi{\k{e}}ba} {and}
  \bibinfo{person}{Jakub~M Tomczak}.} \bibinfo{year}{2015}\natexlab{}.
\newblock \showarticletitle{Boosted SVM with active learning strategy for
  imbalanced data}.
\newblock \bibinfo{journal}{\emph{Soft Computing}} \bibinfo{volume}{19},
  \bibinfo{number}{12} (\bibinfo{year}{2015}), \bibinfo{pages}{3357--3368}.
\newblock


\end{thebibliography}

\end{document}